%% file: paper.tex
\documentclass[]{TEAI}
\usepackage{helvet}
 
\usepackage{natbib} 

\usepackage[toc,page,header]{appendix}
\usepackage{amssymb}
\usepackage{fontawesome}

\usepackage{titletoc}

\usepackage{minitoc}

\usepackage{etoolbox}

\definecolor{lightblue}{RGB}{200, 230, 255}  
\definecolor{headerblue}{RGB}{150, 200, 255} 

\usepackage{pgfplots}
\usepackage[utf8]{inputenc} % allow utf-8 input
\usepackage[T1]{fontenc}    % use 8-bit T1 fonts
\usepackage{url}            % simple URL typesetting
\usepackage{booktabs}       % professional-quality tables
\usepackage{amsfonts}       % blackboard math symbols
\usepackage{nicefrac}       % compact symbols for 1/2, etc.
\usepackage{microtype}      % microtypography
\usepackage{xcolor}         % colors
\usepackage{xspace}
\usepackage{graphicx}
\usepackage{float}
\usepackage{comment}
\usepackage{multirow} % For multi-row cells
\usepackage{amsmath} % For \text command if needed inside math mode\Delta
\usepackage{makecell} % For multi-line cells and better vertical spacing in cells
\usepackage{hyperref}
\usepackage{siunitx}  % For better number alignment (optional but recommended)
\usepackage{tikz}
\usepackage{pgf-pie} % Package for creating pie charts
\usepackage{subcaption}
\usepackage{wrapfig}
\usepackage[export]{adjustbox}

\usepackage{ragged2e}      % for \RaggedRight in tabularx
\usepackage{array}          % For advanced column formatting (like >{\centering\arraybackslash}X)
\usepackage{caption}        % Recommended for figures/tables, but we'll do simple text below images here.
\usepackage{enumitem}
\usepackage{pifont}
\usepackage[hang,flushmargin]{footmisc}

\usepackage{tcolorbox}

\usepackage{tcolorbox} 
\tcbuselibrary{breakable}
\tcbuselibrary{skins} 
\usepackage{tabularx}
\usepackage{listings}

\newcommand*{\system}{DisCo\@\xspace}
\newcommand*{\bench}{DisCoBench\@\xspace}

\newcommand*{\ie}{\textit{i.e.}\@\xspace}

\title{\textsc{DisCo}: World Models with Discrete Camera Motion Control}

\author{
    Hongrui Huang\textsuperscript{1},
    Junke Wang\textsuperscript{1},
    Quanhao Li\textsuperscript{1},  
    Yu-Gang Jiang\textsuperscript{1},
    Zuxuan Wu\textsuperscript{1}
}

\affiliation[1]{\mbox{Fudan University}}

\abstract{
Controllable video world models target interactive world exploration, where models must faithfully execute explicit action commands while preserving visual quality and temporal coherence. However, most existing approaches rely on continuous camera trajectories as action conditions, which often lead to unreliable action following, especially under complex motion sequences.
In this work, we identify action representation entanglement as a key bottleneck in controllable video generation, and show that continuous camera representations lead to high feature similarity across distinct motion patterns, degrading action controllability. Based on this insight, we propose \system, a controllable video world model that conditions generation on a compact set of discrete action primitives to improve action separability. 
We further introduce DisCoBench, a comprehensive benchmark for evaluating the ability of models in short-term, long-horizon, and highly dynamic exploration scenarios. Extensive experiments demonstrate that \system achieves significantly more reliable action following while preserving visual quality.
}

\begin{document}
\maketitle
\renewcommand{\thefootnote}{}
\renewcommand{\thefootnote}{\arabic{footnote}}

\vspace{-1.5em}

\input{section/introduction}

\input{section/relatedwork}
\input{section/method}
\input{section/experiments}
\input{section/conclusion}

\bibliographystyle{plainnat}
\bibliography{main}

\clearpage
\beginappendix
\input{section/appendix}

\end{document}

%% file: section/introduction.tex
\section{Introduction}\label{introduction}

World models learn the dynamics of an environment to forecast future states from past observations and actions~\cite{ha2018world,hafner2019dream,koh2021pathdreamer}. They support model-based decision making by enabling planning and control through imagined rollouts, where an agent evaluates candidate action sequences before execution to improve data efficiency and safety~\cite{bar2025navigation,world4rl}. Recently, advances in video diffusion models have made pixel-space world modeling increasingly practical~\cite{blattmann2023stable,motionctrl,cogvideox,hunyuanvideo,wan,yume,gamecraft,matrixgame,matrixgame2}, with high-fidelity video generators serving as the backbone for high-fidelity simulation.

\begin{figure}
\centering
\includegraphics[width=\textwidth]{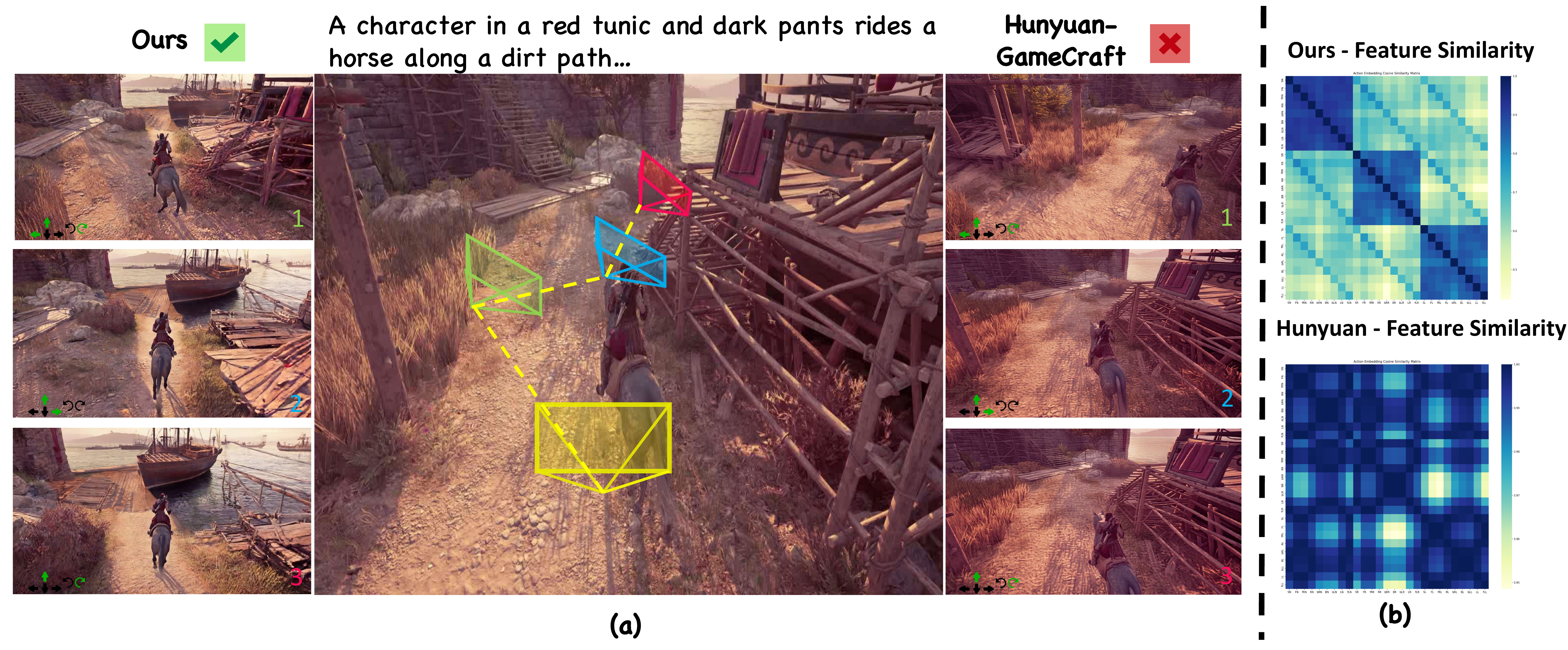}
\caption{
(a) Comparison between the continuous-action-based Hunyuan-GameCraft and our discrete-action-based approach. Our method successfully executes complex action commands, while Hunyuan-GameCraft fails under the same settings.
(b) Heatmaps of pairwise action feature similarity, where darker colors correspond to higher similarity.
}
\label{teaser}
\end{figure}

Despite substantial progress in visual quality and temporal consistency, reliable action controllability remains a critical bottleneck in video world models~\cite{gamecraft,magicworld,matrixgame2,worldplay,astra,motionstream,longvie2,yume}. Most existing approaches~\cite{gamecraft,motionstream,longvie2,worldplay} rely on continuous camera trajectories as action conditions. While effective for coarse motion control, this formulation may collapse distinct action patterns into highly similar conditioning features, thus producing ambiguous guidance and degraded adherence when actions change frequently, as illustrated in Fig.\ref{teaser}.

In this work, we posit that carefully designed discrete action representations provide a stronger foundation for controllable world modeling, as they naturally induce higher separability in the conditional feature space. By partitioning control signals into a compact set of geometrically meaningful action primitives, such representations could offer cleaner supervision for learning complex action compositions and action-conditioned dynamics. Building on this principle, we propose \system, a discrete action-conditioned video diffusion model for highly controllable world modeling.
To support robust controllability under dynamic and complex scenarios, we first curate a camera-controlled video dataset that covers a diverse range of dynamic scenes and camera motion patterns.
Starting from a pretrained video diffusion model, \ie, Wan 2.2~\cite{wan}, \system is fine-tuned for action-conditioned generation with structured motion representations derived from discretized camera controls. To enable efficient long-horizon video generation, we distill \system into a four-steps causal model, incorporating windowed attention and attention sinks with rolling KV caches to preserve temporal coherence and long-term consistency.

To comprehensively evaluate the proposed approach, we further introduce \bench, a carefully curated benchmark that assesses controllable video generation across various scenarios. Unlike existing evaluations that primarily focus on visual and temporal quality, \bench systematically assesses the ability of models to (i) accurately follow complex and compositional action commands in various scenes, (ii) maintain temporal coherence and visual consistency over long horizons, and (iii) handle scenarios involving highly dynamic elements. The results on \bench show that \system exhibits superior controllability on challenging camera conditions, achieving an action following score of 0.633 compared to 0.588 for Matrix-Game2.

%% file: section/relatedwork.tex
\section{Related Work}
\subsection{Camera Controllable Video Generation}
Camera control plays a central role in interactive video generation as it directly determines the viewpoint dynamics. Existing methods~\cite{motionctrl, cameractrl, wan} typically encode explicit camera intrinsics and extrinsics as control signals, which are then injected into diffusion models via ControlNet-style~\cite{controlnet} adapters. Building on this paradigm, Gen3C~\cite{gen3c} and AC3D~\cite{ac3d} further incorporate 3D geometric priors to improve scene consistency, while ReCamMaster~\cite{recammaster} and ReCapture~\cite{recapture} focus on novel view synthesis by conditioning on relative camera pose trajectories. Instead of using adapters, MagicWorld~\cite{magicworld} leverages intermediate 3D representations, projecting reconstructed 3D structure to the target view and completes unseen regions via generative inpainting.
Despite their impressive visual quality, these approaches have two key limitations:
(i) their controllability normally degrades when faced with complicated actions.
(ii) raw numerical camera parameters are cumbersome and unintuitive for human interaction, limiting practical usability in interactive settings.

\subsection{Auto-regressive Video Generation}
Autoregressive generation~\cite{videopoet,sun2024autoregressive,wang2025simplear} has emerged as a powerful paradigm for video generation, typically utilizing next-token prediction to capture temporal dynamics~\cite{videogpt,wang2024omnitokenizer,wang2026omnigen}. Some works have explored hybrid AR-diffusion architectures~\cite{diffusion-forcing, streamt2v,pyramidalflow,arlon,lct}, integrating efficiency of AR with high quality of diffusion. Most recently, Self-forcing~\cite{self-foring}, and its following works~\cite{self-forcing++,rolling-forcing,reward-forcing} address the train-test gap by simulating inference conditions during training, and distill slow, multi-step teacher model into a fast, few-step AR student. Inspired by this, we develop \system and enhance it on controllability and consistency.

\subsection{Video World Models}
Video world models~\cite{gamecraft,yume,matrixgame,matrixgame2,gamefactory,ye2025yan,li2025magicmotion,motionstream,longvie2} have recently made significant strides by leveraging large-scale transformer architectures. Conditioning generation on actions or camera trajectories, these systems learn to synthesize high-fidelity, temporally coherent environments for interactive exploration. 
However, their controllability often degrades under complex and compositional action sequences, and the underlying causes of this degradation remain largely unexplored. 
Both discrete-action-based models ~\cite{yume,matrixgame2} and continuous-action-based models ~\cite{gamecraft,worldplay,motionstream,longvie2} primarily treat action conditioning as a system-level design choice, with limited analysis of its representation-level implications.

%% file: section/method.tex
\section{Method}
\subsection{Overview}
Our goal is to build a video world model that can reliably follow complex camera motion commands. 
Formally, given an initial frame $x_0$, a text prompt $\tau$, and an action sequence ${a}_{seq}$, 

we aim to generate a video whose dynamics align with both semantic contents and specified actions. 
To this end, we adopt a pretrained image-to-video diffusion model~\cite{wan} as the backbone,
which synthesizes videos conditioned on $x_0$ and $\tau$.

The overall pipeline consists of three stages:
(i) an \textbf{Action Discretization} stage that maps continuous camera trajectories $\mathbf{T}$ to discrete action sequences ${a}_{seq}$ ,
(ii) an \textbf{Action-Controllable Teacher Model} that generates videos conditioned on $a_{seq}$, and
(iii) an \textbf{Action-Guided Distillation} procedure that distills the teacher model into an efficient autoregressive student for long-horizon generation.

\subsection{Action Discretization}\label{sec:discretization}
Given a camera trajectory $\mathbf{T} = \{\mathbf{T}_t\}_{t=0}^{T}$,
where each $\mathbf{T}_t$ is a $4\times4$ camera-to-world pose for the $t$-th frame,
we map it into a discrete sequence $a_{seq}$ with the following steps.

\textbf{Temporal partitioning.}
We first divide $\mathbf{T}$ into a sequence of short temporal segments to capture locally coherent camera motions. The initial pose $\mathbf{T}_0$ is treated as a reference pose, and the remaining poses are partitioned into consecutive segments of fixed length $k=4$. As a result, the trajectory is divided into $N$ segments, with each summarizing camera motion over a short temporal interval and serving as the basic unit for action extraction.

\textbf{Relative pose estimation.}

Let $\mathbf{T}_{k-1}$ and $\mathbf{T}_{k}$ denote the terminal poses of two adjacent segments, we then calculate their relative transformation to represent the camera action over the segment:
\begin{equation}
\Delta \mathbf{T} = \mathbf{T}_{k}\mathbf{T}_{k-1}^{-1} =
\begin{bmatrix}
\Delta\mathbf{R} & \Delta\mathbf{t} \\
\mathbf{0}^\top & 1
\end{bmatrix}
\end{equation}

where $\Delta \mathbf{T} \in SE(3)$ is a $4\times4$ homogeneous transformation matrix, $\Delta\mathbf{R} \in SO(3)$ is the rotation matrix and $\Delta\mathbf{t} = \left[dx, dy, dz\right]^\top$ is the translation vector. 

\textbf{Quantization and primitive mapping.} 

We follow DeepVerse~\cite{deepverse} to decompose $\Delta \mathbf{T}$ into two decoupled motion primitives through a threshold-based quantization process. Specifically, the translation vector $\Delta\mathbf{t} = \left[dx, dy, dz\right]^\top$ and rotation matrix $\Delta\mathbf{R}$ are separately mapped to a set of symbolic labels in the following manner:
\begin{itemize}
    \item \textbf{Translation ($N_t=9$ types):} we compute the horizontal displacement $d = \sqrt{dx^2 + dz^2}$. If $d$ is below a predefined threshold $\tau_t$, the motion is assigned a \textit{Stationary} label $s_0$. Otherwise, the translation is categorized into one of eight directional bins based on the quantized polar angle $\phi = \operatorname{arctan2}\left(dx, dz\right)$, representing: \textit{Forward} ($\text{F}$), \textit{Backward} ($\text{B}$), \textit{Left} ($\text{L}$), \textit{Right} ($\text{R}$), and their diagonal counterparts (\textit{fL, fR, bL, bR}).
        
    \item \textbf{Rotation ($N_r=3$ types):} we extract the yaw angle $\psi$ from $\Delta\mathbf{R}$. The rotation is mapped to a discrete state $r \in \{\text{N, L, R}\}$, where $\text{N}$ represents a \textit{Neutral} state ($|\psi| \le \tau_r$), $\text{L}$ (Left) and $\text{R}$ (Right) represent counter-clockwise and clockwise rotations exceeding the threshold $\tau_r$, respectively.
\end{itemize}

Combining these primitives, each camera action is represented as an action pair $a_k = \left(i_k, j_k\right)$, defined in a discrete action space $\mathcal{A}$ of size $9 \times 3 = 27$, where $i_k \in \{0, \dots, 8\}$ and $j_k \in \{0, \dots, 2\}$ index the corresponding translation category (e.g., S, F, fL) and rotation category (e.g., N, L, R) respectively.
In this way, we convert the continuous camera trajectory $\mathbf{T}$ into a discrete action sequence ${a_{seq}} = \{a_1, a_2, \dots, a_{N}\}$.

\begin{figure*}[!t]
\vspace{-1em}
\begin{center}
\centerline{\includegraphics[width=\textwidth]{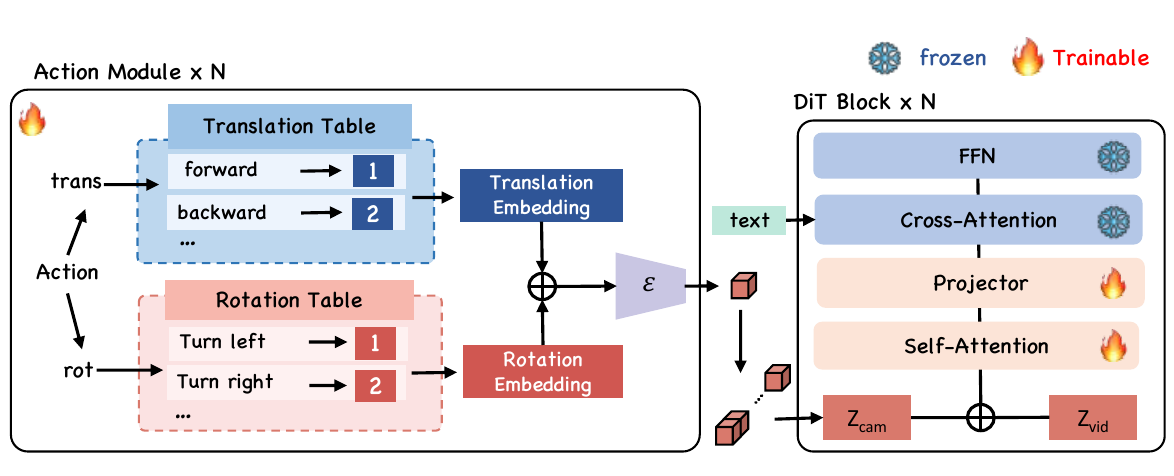}}
\caption{Architecture of the action-controllable teacher model. Each discretized action is decomposed into translation and rotation indices, which retrieve embeddings from separate lookup tables. The embeddings are projected and concatenated to form the action feature, and then fused with video latents via element-wise addition.}
\label{model}
\end{center}
\vspace{-2em} 
\end{figure*}

\subsection{Action-Controllable Teacher Model}

To supply our image-to-video backbone~\cite{wan} with action controllability,
we introduce an action module that maps $a_{\text{seq}}$ into a high-dimensional latent space and then fuse it with video latents, as illustrated in Fig.~\ref{model}.

Specifically, we construct two learnable embedding tables, $\mathbf{E}_{trans} \in \mathbb{R}^{9 \times d}$ and $\mathbf{E}_{rot} \in \mathbb{R}^{3 \times d}$ to obtain the action embedding $e_{k}\in \mathbb{R}^d$ based on $a_k = \left(i_k, j_k\right)$: 

\begin{equation}
    \mathbf{e}_{k} = 
    \begin{cases} 
    \mathbf{e}_{start}, & k = 0 \\
    \text{Proj}\left(\mathbf{E}_{trans}\left[i_k\right] {+} \mathbf{E}_{rot}\left[j_k\right]\right), & k > 0
    \end{cases}
\end{equation}
where $\mathbf{e}_{start} \in \mathbb{R}^d$ is a learnable parameter representing the initial state. $\mathbf{E}[\cdot]$ denotes a lookup operation that retrieves the $i$-th or $j$-th row from the respective embedding tables, indexed by the action categories.

$\text{Proj}(\cdot)$ is a linear projection layer, initialized as zeros to stabilize training by gradually introducing
action-conditioned signals.

The resulting action embedding can be represented as $e_{act} = \{e_0, e_1, \dots, e_N \}$.

The action embedding $e_{act}$ is integrated into the video features $\mathbf{f}_{vid} \in \mathbb{R}^{b \times (t \times h \times w) \times d}$ through a structured expansion and fusion process.
Specifically, the sequence of action embeddings $\{\mathbf{e}_k\}_{k=0}^{t-1}$ is stacked along the temporal axis to form a temporal action representation
$\mathbf{f}_{act}^{temp} \in \mathbb{R}^{b \times t \times d}$.
To match the resolution of the video features, each temporal embedding is replicated $h \times w$ times and flattened to the token dimension, resulting in
$\mathbf{f}_{act} \in \mathbb{R}^{b \times (t \times h \times w) \times d}$.
This expanded action feature is combined with the video features via element-wise addition to get the fused feature:
\begin{equation}
\mathbf{f}_{fuse} = \mathbf{f}_{vid} + \mathbf{f}_{act}.
\end{equation}
Following ReCamMaster~\cite{recammaster}, $\mathbf{f}_{fuse}$ are further processed by a self-attention layer together with an additional linear projection layer, enabling the action-conditioned signals to modulate the video representation more effectively.

The model is then optimized via the following objective:
\begin{equation}
    \mathcal{L} = \mathbb{E}_{z, \epsilon, t, \mathbf{f}_{act}, \tau, x_0} \left[ \| \epsilon - \epsilon_\theta\left(z_t, t, \mathbf{f}_{act}, \tau, x_0\right) \|^2 \right]
\end{equation}
where $t$ denotes the diffusion timestep, $\tau$ is the text prompt, $x_0$ is the conditional frame, $\epsilon$ is the added noise, and $\epsilon_\theta$ is the denoising network~\cite{wan}.

\subsection{Action-Guided Distillation}
To enable efficient long video generation, we distill the teacher model $G_{teacher}$ into a four-step autoregressive student $G_{student}$ using self-forcing strategy~\cite{self-foring}, which consists of two stages: ode initialization and DMD~\cite{dmd2}-based training.

\textbf{Attention pattern.}
During both training and inference, we adopt a truncated causal attention mechanism consisting of a fixed-size local window and a set of sink tokens\cite{longlive, streamllm, motionstream}. Concretely, each token attends to (i) all preceding sink tokens and (ii) a sliding window of recent tokens. We denote the number of sink tokens as $S$ and the local window size as $W$.

\textbf{Ode initialization.}
We first construct a dataset $\mathcal{D}$ comprising ODE trajectories $\{\mathbf{z}_t^i\}_{i=1}^N$ generated by $G_{teacher}$, where the timesteps $t$ are sampled  from a 4-step subset of $[0, T]$. 

Following the initialization protocol from CausVid~\cite{causvid}, the input of $N$ noisy latents is partitioned into $L$ non-overlapping chunks $\{\mathbf{z}_{t^i}^i\}_{i=1}^L$ with independent timesteps, 

and we adopt a causal attention mask $M$ that permits the attention pattern we use. 

Specifically, the mask $M \in \{0,1\}^{N \times N}$ is defined as:
\begin{equation}
M\left[m,n\right]
=
\begin{cases}
1, & n < S \;\wedge\; n \le m, \\[2pt]
1, & S \le n \le m \;\wedge\; n \ge m - W, \\[2pt]
0, & \text{otherwise}.
\end{cases}
\end{equation}

For each chunk $i$, the attention mechanism is constrained to $M$. The final optimizing objective can be formulated as: 
\begin{equation}
    \mathcal{L}_{\text{ode}} = \frac{1}{L} \sum_{i=1}^{L} \mathbb{E}_{\mathbf{z}, t^i} \left\| G_{student} \left( \mathbf{z}_{t^i}^i, c^i, t^i, M \right) - \mathbf{z}_0^i \right\|^2
\end{equation}
where $c^i$ includes the text $\tau$ and action conditions $a_i$.

\textbf{DMD-based self-forcing training.}
Subsequently, we perform self-forcing style distillation leveraging distribution matching distillation (DMD)~\cite{dmd2}. Unlike traditional teacher-forcing, $G_{student}$ autoregressively produces video chunks $\{{\mathbf{z}}_0^i\}_{i=1}^L$ conditioned on its own previous outputs, mitigating the training-inference gap. To balance long-range stability and computational efficiency, we implement a rolling KV-cache mechanism with an anchored attention sink.
Specifically, for each chunk $i$, the generative process is conditioned on the corresponding action signals $a_i$, text condition $\tau$, and contexts $\mathcal{C}_i$:
\begin{equation}
\mathcal{C}_i = \{ \mathbf{z_0^j} \mid 0 \le j < S \text{ or } \max\left(0, i - W\right) \le j < i \}
\cup \{\mathbf{z_t^{i}}\}
\end{equation}

The generator is optimized by minimizing the approximate KL-divergence between the synthesized distribution $p_\theta$ and the real data distribution $p_{\text{real}}$. An auxiliary distribution $p_{\text{fake}}$ is utilized to estimate $p_\theta$, and the gradient for the generator parameters $\theta$ can be formulated as:

\begin{equation}
    \nabla_\theta \mathcal{L}_{\text{DMD}} = -\mathbb{E}_{t} \left(\nabla_\theta \text{KL}\left(p_{\text{fake},t} \| p_{\text{real},t}\right) \right)
\end{equation} 

%% file: section/experiments.tex
\section{Experiments} \label{Experiments}
\textbf{Training data.}
To enable our model to handle scenarios with highly dynamic contents, we select around 10k high-quality clips from OmniWorld~\cite{omniworld}, and re-annotate their camera poses with VIPE~\cite{vipe}. 

However, We observe a long-tail distribution in action types within this subset, where forward-dominant sequences account for the majority of samples.
To relieve this imbalance and improve generalization, we further augment it with synthetic data from a UE simulator~\cite{context_as_mem}. The final dataset comprises approximately 20k videos of 81 frames, featuring a rich variety of camera motions. 

\textbf{Implementation Details.}
We build \system based on the Wan2.2-5B image-to-video generation model~\cite{wan}. 
For action discretization, the translation threshold $\tau_t$ is 0.1 and the rotation threshold $\tau_r$ is 1.
For the ode initialization, we use $G_{teacher}$ to generate 4k ode pairs with a balanced action distribution as our training data. In the self-forcing distillation stage, $G_{student}$ generates videos chunk-wisely, it takes the first conditional image as the first chunk and splits the remaining video with the chunk size$=$4. In both ode initialization and self-forcing stage, the sink size is 1 and the local attention window size is 8. More details are in Appendix~\ref{More Experimental details}

\textbf{DisCoBench.}
To comprehensively evaluate our model, we introduce DisCoBench, which consists of 3 subsets to assess generation capabilities from different perspectives:
\begin{itemize}
\item \textbf{DisCo-ActionBench}: This subset contains 300 initial images from diverse scenes paired with predefined complex action sequences. Each sequence contains at least two kinds of actions from our unified action set $\mathcal{A}$. This subset evaluates the controllability and short video generation ability of the model. More details about DisCo-ActionBench can be found in Appendix.\ref{appendix: A}. 
\item \textbf{DisCo-DynamicBench}: This subset includes 200 dynamic third-person exploration videos to specifically measure the ability of models to generate videos with highly dynamic elements.
\item \textbf{DisCo-LongBench}: This subset is an extension of the DisCo-ActionBench with prolonged action sequences to evaluate long-horizon stability. Specifically, we extend each action for 3 times, supporting evaluation for long videos with at least 241 frames generation
\end{itemize}

\subsection{Evaluation on DisCo-Bench}
\textbf{Evaluation on DisCo-ActionBench.} In this section, we mainly evaluate the action-following performance and the visual-temporal quality of the generated videos on DisCo-ActionBench.

For the temporal coherence and visual fidelity evaluation, we adopt the widely used VBench++ protocol~\cite{vbench++}, including Subject Consistency, Background Consistency, Motion Smoothness, Aesthetic Quality, and Imaging Quality.
For the evaluation of action following ability, we use a carefully designed metric,  \ie, Action Score. Specifically, we utilize VIPE ~\cite{vipe} to obtain the c2w poses from the generated videos, and then transform the resulting camera trajectories into discrete action sequences with the discretization approach described in \ref{sec:discretization}. The Action Score is defined as the proportion of actions that exactly match the ground-truth actions across all samples. 

We compare our model against two categories of baselines: (1) Continuous-actions-based models: CameraCtrl~\cite{cameractrl}, MotionCtrl~\cite{motionctrl}, WanX-Cam~\cite{wan},
and Hunyuan-GameCraft~\cite{gamecraft}.
The CameraCtrl and MotionCtrl adopts the image-to-video
SVD version. WanX-Cam corresponds to the implementation of VideoX-Fun. (2) Discrete action based models:
Matrix-Game2\cite{matrixgame2}.
As shown in Tab.\ref{ShortExploreGen}, our method achieves the highest action score while maintaining superior visual and temporal quality.
\begin{table*}[ht]
\vspace{-0.5em}
  \caption{Comparison on DisCo-ActionBench. \textbf{Bold} and \underline{underlined} indicate the best and second-best results respectively.}
  \label{ShortExploreGen}
  \begin{center}
    \begin{small}
        \setlength{\tabcolsep}{4.5pt}
        
        \begin{tabular}{clccccccr}
          \toprule
           &Model  & \makecell{Subject \\ Consistency} & \makecell{Background \\ Consistency} & \makecell{Motion \\ Smoothness} & \makecell{Aesthetic \\ Quality} & \makecell{Imaging \\ Quality} & \makecell{Action \\ Score}      \\   
          \midrule
          \multirow{4}{*}{Continuous} 
          &CameraCtrl    & 0.8265   & 0.9202    & 0.9604    & 0.4536    & 0.5358    & 0.5731        \\
          &MotionCtrl    & \underline{0.8709}   & 0.9294    & 0.9797    & 0.4608    & 0.5618    & 0.3636        \\
          &WanX-Cam      & 0.8139   & 0.8977    & 0.9764    & 0.5070    & 0.6351    & 0.5143    &    \\
          &GameCraft     & 0.8559   & 0.9169    & 0.9757    & 0.5074    & \textbf{0.6792}    & 0.2736        \\
          \midrule
          \multirow{2}{*}{Discrete} 
          &Matrix-Game2  & 0.8538   & \underline{0.9333}    & \underline{0.9824}    & \underline{0.5148}    & \underline{0.6474}    & \underline{0.5880} \\
          &Ours          & \textbf{0.8869}   & \textbf{0.9353}    & \textbf{0.9842}    & \textbf{0.5308}    & 0.6420    & \textbf{0.6332}        \\
          \bottomrule
        \end{tabular}
        
    \end{small}
  \end{center}
  \vspace{-0.5em}
\end{table*}

\textbf{Evaluation on DisCo-DynamicBench.}
The ability to generate exploration videos with dynamic elements is crucial for interactive experiences. To evaluate this capability, we construct DisCo-DynamicBench, a benchmark tailored for assessing dynamic exploration video generation. We adopt two metrics for evaluation: Fréchet Video Distance (FVD) ~\cite{fvd} and VQA-Score ~\cite{vqa}.

FVD measures the distributional similarity between generated videos and reference exploration videos, where a lower score indicates closer alignment with the target video distribution. VQA-Score evaluates whether the generated videos correctly follow the textual prompts. 
For instance, if a prompt requires a character to perform a running action but the generated video only exhibits camera motion with a static character, the resulting VQA-Score will be penalized. 
We use Qwen2.5-VL-7B~\cite{qwen2.5} as the underlying vision-language model for VQA scoring.
The results are illustrated in Tab.~\ref{DynamicExploreGen}. \system achieves the best FVD score and competitive VQA-Score performance, demonstrating its effectiveness in generating exploration videos with dynamic elements. More details on the effectiveness of VQA-Score are provided in Appendix~\ref{appendix: A}.
\begin{table}[ht]
  \caption{Comparison on the \textbf{DisCo-DynamicBench}}
  \label{DynamicExploreGen}
  \begin{center}
    \begin{small}
        \setlength{\tabcolsep}{16pt}
        \begin{tabular}{lccr}
          \toprule
          Model  & FVD  & VQAScore  \\   
          \midrule
          CameraCtrl    & 470.22   & 0.3321   \\
          MotionCtrl    & 481.19   & 0.4126   \\
          WanX-Cam      & \underline{285.95}   & \textbf{0.5637}   \\
          GameCraft     & 374.06   & 0.4033   \\
          Matrix-Game2  & 380.28   & 0.3884   \\
          Ours          & \textbf{220.51}   & \underline{0.5448}     \\
          \bottomrule
        \end{tabular}
    \end{small}
  \end{center}
\end{table}

\textbf{Evaluation on DisCo-LongBench.}
To evaluate the long video generation ability of \system, we compare it with two controllable long video generation models: Matrix-Game2~\cite{matrixgame2} and Hunyuan-GameCraft~\cite{gamecraft}. As shown in Tab.\ref{LongExploreGen}, \system outperforms in consistency, motion smoothness and aesthetic quality, while achieving competitive imaging quality.
\begin{table*}[ht]
  \caption{Comparison with relative baselines on the \textbf{DisCo-LongBench}}
  \label{LongExploreGen}
  \begin{center}
    \begin{small}
        \begin{adjustbox}{width=\textwidth, center}
        \begin{tabular}{lccccccr}
          \toprule
          Model  & Subject Consistency & Background Consistency & Motion Smoothness & Aesthetic Quality & Imaging Quality  \\   
          \midrule
          GameCraft     & 0.7615   & 0.8842    & 0.9740    & 0.4758    & 0.6441                \\
          Matrix-Game2  & 0.7593   & 0.8969    & 0.9814    & 0.4883    & \textbf{0.6474}      \\
          Ours          & \textbf{0.8138}   & \textbf{0.9103}    & \textbf{0.9846}    & \textbf{0.4938}    & 0.6372 \\
          \bottomrule
        \end{tabular}
        \end{adjustbox}
    \end{small}
  \end{center}
\end{table*}
\begin{figure*}[ht]
  \vspace{-1em}
  \begin{center}
    \centerline{\includegraphics[width=\textwidth]{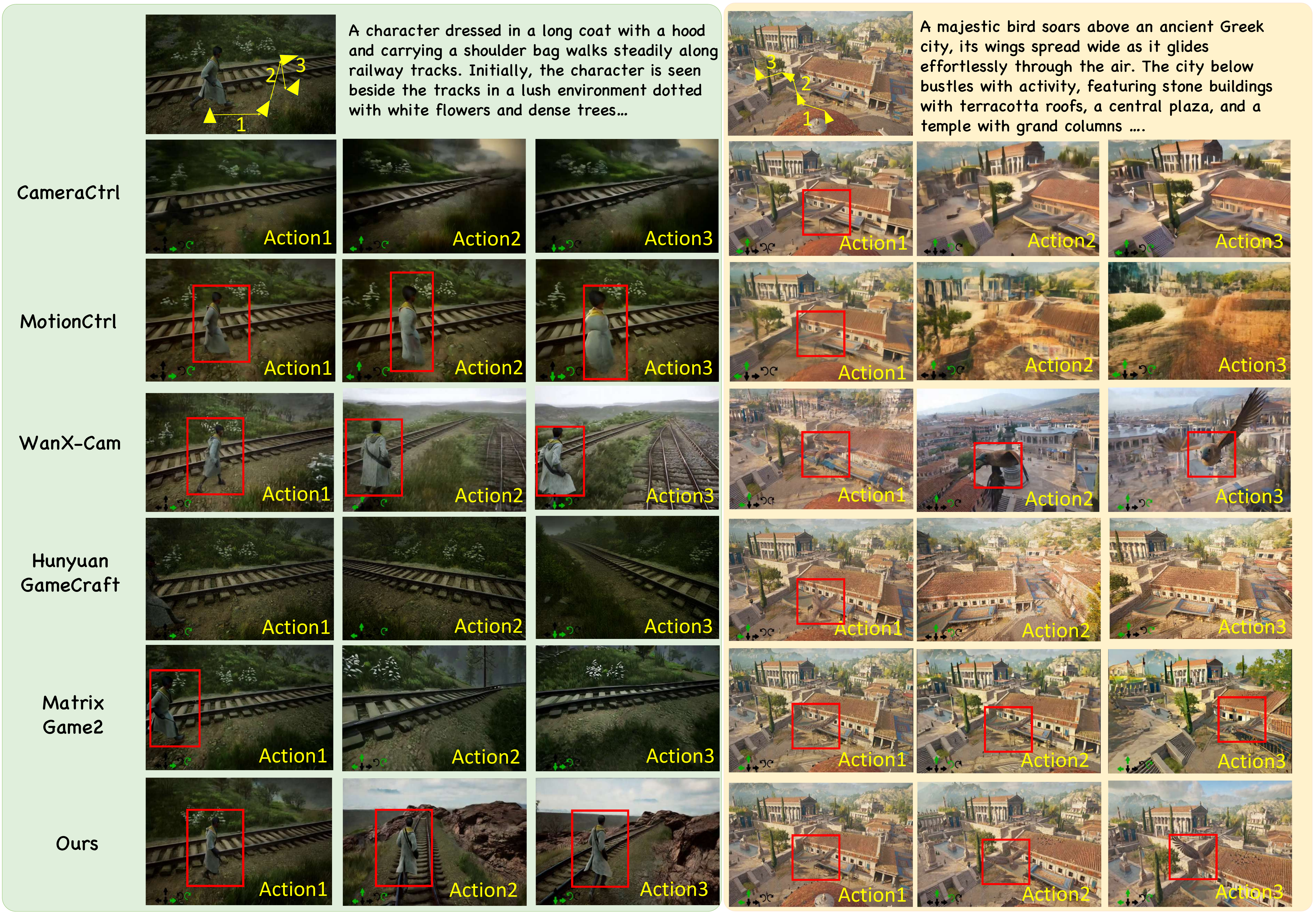}}
    \caption{
    Comparison with baselines on complicated camera actions. The red box means the dynamic element in the generated video. \system is the only model with both great controllability and the ability to generate videos with dynamic elements.
    }
    \label{comparison_short}
      \vspace{-2.5em}
  \end{center}
\end{figure*}

\textbf{Qualitative Comparison.}
As shown in Fig.~\ref{comparison_short}, MotionCtrl and Hunyuan-GameCraft struggle with both action following and dynamic content generation. CameraCtrl and Matrix-Game2 follow complex action commands but fails to generate the required dynamic elements. WanX-cam exhibits noticeable visual inconsistencies. Comparatively, our method achieves accurate action following while maintaining stable generation of dynamic elements. More demos are shown in Appendix~\ref{appendix: demo}.

\subsection{Continuous versus Discrete Actions} \label{sec: 4.1}
As discussed in Sec.\ref{introduction}, limited separability of action features leads to ambiguous guidance and degraded controllability in continuous-action-based models, whereas carefully designed discrete action features alleviate this issue. In this subsection, we empirically validate this by analyzing the similarity of action features and the resulting controllability.

For a direct analysis of action features, we focus on adapter-based models that explicitly extract action features before fusion with visual features. Based on this criterion, we select representative continuous-action-based baselines, including CameraCtrl~\cite{cameractrl}, WanX-Cam~\cite{wan}, and Hunyuan-Gamecraft~\cite{gamecraft} for further comparison.

\textbf{Action feature similarity.}
To assess whether our method produces more discriminative conditioning features, we analyze the cosine similarity of action features across different motion patterns.
Specifically, we define the similarity score as the average pairwise cosine similarity:
\begin{equation}
\mathcal{S}
=
\frac{1}{N(N-1)}
\sum_{i \neq j}
\frac{\mathbf{f}_i^\top \mathbf{f}_j}
{\|\mathbf{f}_i\|_2 \, \|\mathbf{f}_j\|_2}.
\end{equation}
where $\mathbf{f}_i$ denotes the action feature extracted from the action encoders. Each action pattern is sampled from the unified action set $\mathcal{A}$ mentioned in \ref{sec:discretization}. For models conditioned on continuous camera controls, c2w matrices are used as conditioning inputs, of which different transformations correspond to different discretized actions in $\mathcal{A}$.
\begin{wrapfigure}{r}{0.45\textwidth} 
  \centering
  \vspace{-10pt} 
  \includegraphics[width=0.43\textwidth]{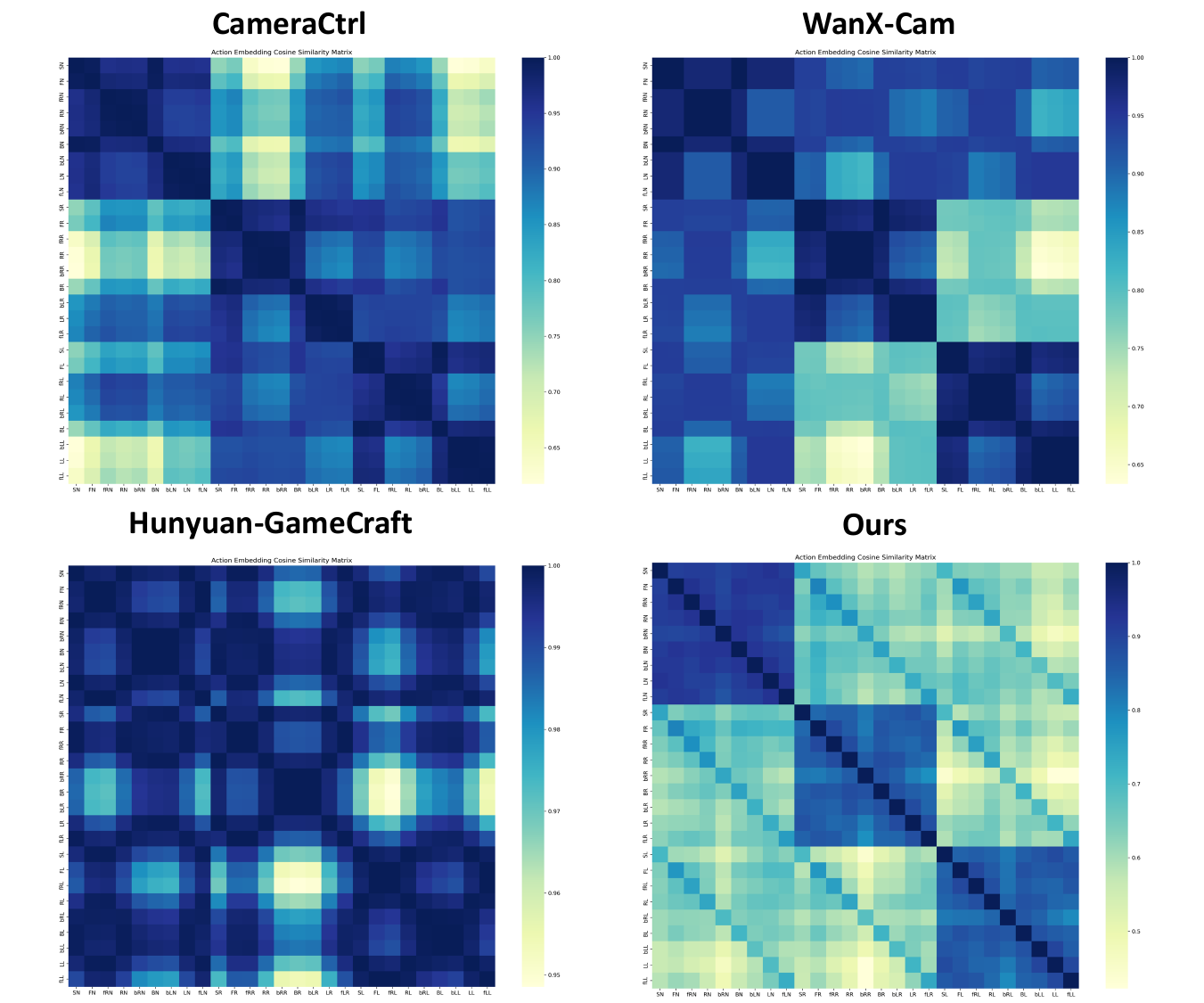}
  \vspace{-5pt} 
  \caption{Heatmaps of action feature similarity for different adapter-based models. Darker colors indicate higher cosine similarity.}
  \label{heatmap}
  \vspace{-10pt} 
\end{wrapfigure}
As illustrated in Fig.~\ref{heatmap}, our method produces action features with more regular and interpretable similarity structures compared to continuous-action-based methods.
Specifically, actions sharing the same rotation pattern form coherent clusters, and actions with the same translation pattern also exhibit certain similarity.
In contrast, continuous-action-based models produce highly entangled and less interpretable feature representations, resulting in uniformly higher similarity scores, as quantitatively reflected in Tab.~\ref{feature_similarity}.

\textbf{Controllability evaluation.} As illustrated in Tab.\ref{feature_similarity}, 
Our Method achieves the best action score, while continuous-action-based models with entangled representations exhibit weaker controllability. These findings provide empirical support for our claim that preserving sufficient action separability in the conditioning feature space is critical for reliable action control.
We further substantiate this with ablation studies in Sec.~\ref{ablations}.
\begin{table}[ht]
  \caption{Comparison with adapter-based models using continuous action condition}
  \label{feature_similarity}
  \begin{center}
    \begin{small}
        \setlength{\tabcolsep}{16pt}
        \begin{tabular}{lccr}
          \toprule
          Model  & \makecell{Similarity \\ Score}  & \makecell{Action \\ Score} \\   
          \midrule
          CameraCtrl    & 0.8826    & 0.5731   \\
          WanX-Cam      & 0.8953    & 0.5143   \\
          GameCraft     & 0.9913    & 0.2736   \\
          Ours          & \textbf{0.7029}  & \textbf{0.6332}     \\
          \bottomrule
        \end{tabular}
    \end{small}
  \end{center}
 
\end{table}
\subsection{Ablations} \label{ablations}
\textbf{Discrete Action Representation.}
As analyzed in Sec.~\ref{sec: 4.1}, continuous action representations are associated with higher action feature similarity and degraded controllability. 
To isolate the effect of action representation from other factors, we conduct an ablation by replacing the discrete action embedding in \system with a continuous Plücker embedding derived from camera-to-world matrices, while keeping all other architectures and training settings unchanged.
\begin{table}[htbp]
  \caption{Ablation on continuous and discrete representation}
  \label{action_ablation}
  \begin{center}
    \begin{small}
        \setlength{\tabcolsep}{16pt}
        \begin{tabular}{lccr}
          \toprule
             Action           & \makecell{Similarity \\ Score}      & \makecell{Action \\ Score} \\   
          \midrule
              continuous      & 0.9201                  & 0.5563   \\
              discrete(ours)  & \textbf{0.7029}    & \textbf{0.6332}   \\
          \bottomrule
        \end{tabular}
    \end{small}
  \end{center}
\end{table}

As shown in Tab.~\ref{action_ablation}, conditioning on continuous Plücker embeddings results in significantly higher feature similarity and lower action-following accuracy compared to the discrete representation. 
This ablation provides direct evidence that preserving action separability in the conditioning feature space is critical for effective action control.

\textbf{Attention Sink.}
The attention sink is designed to enhance consistency in long video generation. In our setting, we adopt a sink size of 1(take the first frame as the global sink), which aligns with the image-conditioning paradigm used in image-to-video generation. To further investigate the impact of the attention sink, we conduct an ablation study by training models without the attention sink, and evaluate their consistency performance on DisCo-LongBench.

\begin{wrapfigure}{r}{0.5\columnwidth}
  \centering
  \vspace{-10pt} 
  \includegraphics[width=0.45\textwidth]{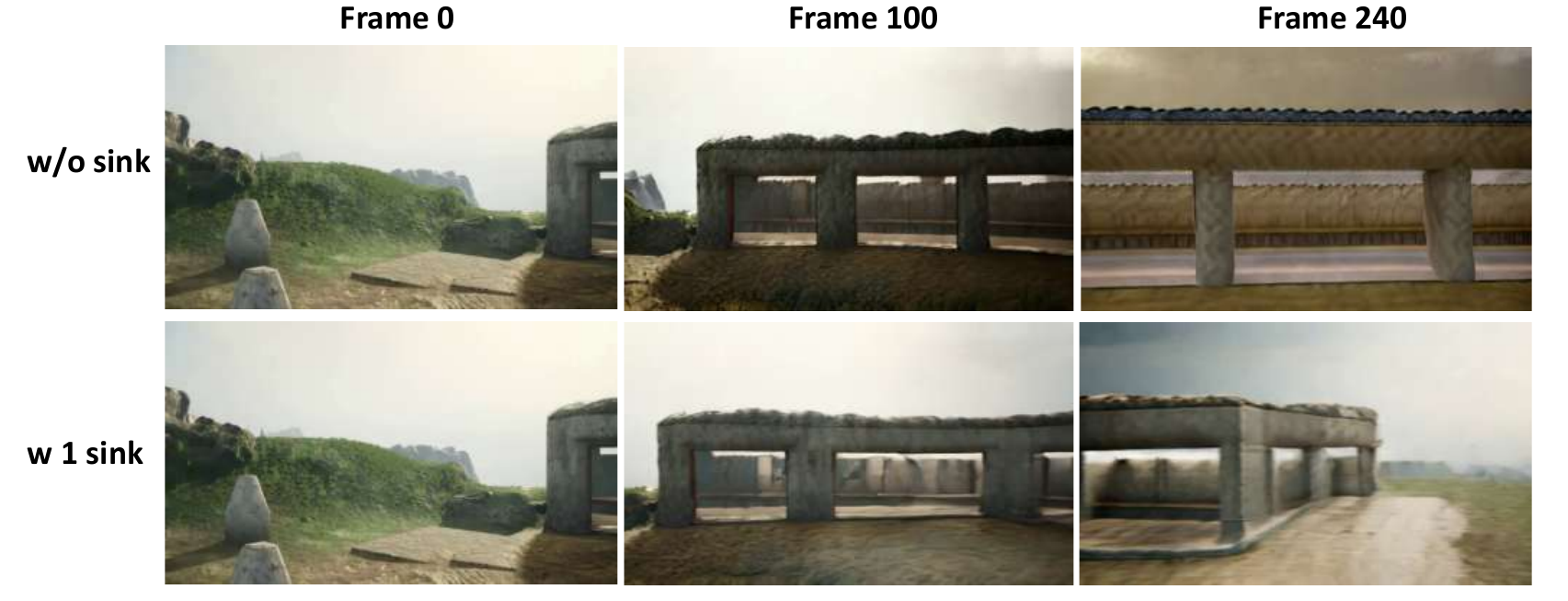}
  \vspace{-5pt} 
  \caption{Ablation on long video generation with and without attention sink.}
  \label{sink}
  \vspace{-8pt} 
\end{wrapfigure}

The results are summarized in Tab.~\ref{sink_ablation}, with qualitative comparisons shown in Fig.~\ref{sink}. Both demonstrate that introducing an attention sink improves the long-horizon consistency. 

\begin{table}[ht]
  \caption{Ablation on attention sink, test on the DisCo-LongBench}
  \label{sink_ablation}
  \begin{center}
    \begin{small}
        \setlength{\tabcolsep}{16pt}
        \begin{tabular}{lccr}
          \toprule
            \makecell{w/wo \\ sink}   & \makecell{Subject \\ Consistency}  & \makecell{Background \\ Consistency} \\   
          \midrule
              w/o sink      & 0.8005    & 0.9037   \\
              w 1 sink      & \textbf{0.8138}    & \textbf{0.9103}   \\
          \bottomrule
        \end{tabular}
    \end{small}
  \end{center}
\end{table}
\textbf{Data Ablation.}
We further conduct ablation experiments to analyze the contribution of different components of our training data. By default, we train \system on a 1:1 mixture of motion-intensive exploration videos ($D$) and camera-stabilized synthetic scenes ($S$); the former features complex actions and dynamic elements, while the latter prioritizes smooth motion and static environment. To isolate their respective effects, we consider two alternative settings: (i) training exclusively on $D$, and (ii) training solely on $S$.

As shown in Tab.~\ref{data_ablation}, training with $D$ enables the model to generate highly dynamic exploration videos. However, the long-tail distribution of its camera actions and the high complexity of such data make it challenging for the model to reliably learn the correct motion patterns. In contrast, $S$ provides more structured and balanced motion supervision, and substantially improves action-following performance. Nevertheless, in setting (ii), the model exhibits degraded FVD and VQA-Score, indicating limited capability in generating videos with rich dynamic elements. These results indicate that combining $D$ and $S$ is crucial for achieving both action controllability and dynamic scene generation.
\begin{table}[ht]
  \caption{Ablation on Dynamic and synthetic ratio}
  \label{data_ablation}
  \begin{center}
    \begin{small}
        \setlength{\tabcolsep}{10pt}
        \begin{tabular}{lcccr}
          \toprule
            D:S         & FVD       & VQAScore    & Action Score  \\   
          \midrule
            1:0         & \textbf{198.96}    & \textbf{0.5779}      & 0.4828        \\
            0:1         & 699.83    & 0.5237      & \underline{0.6005}        \\
            1:1(ours)   & \underline{220.51}    & \underline{0.5448}      & \textbf{0.6332}        \\ 
          \bottomrule
        \end{tabular}
    \end{small}
  \end{center}
  \vspace{-10pt}
\end{table}

%% file: section/conclusion.tex
\section{Conclusion}
This paper presents \system, a video world model with discrete camera motion control. By discretizing camera trajectories into a compact set of predefined action types, DisCo alleviates action representation entanglement, enabling more reliable action following under complex and compositional motion sequences. Combined with self-forcing distillation and attention sinks, \system supports efficient controllable long video generation. 
To facilitate comprehensive evaluation of model performance, we also introduce \bench, which assesses complementary aspects of controllable video generation with different subsets. Extensive experiments demonstrate that \system consistently outperforms prior methods in action controllability, while preserving visual consistency and temporal coherence in dynamic and long-horizon video generation.

%% file: section/appendix.tex
\section{More Experimental details}\label{More Experimental details}
All training stages are conducted on 16 NVIDIA H20 GPUs using the AdamW~\cite{adam} optimizer. 

\textbf{Teacher Model.} 
The teacher model is trained for approximately 10k steps with a batch size of 64 and a learning rate of $1\times10^{-5}$.

\textbf{ODE Initialization Stage.} 
The ODE initialization stage is trained for approximately 8k steps with a batch size of 64 and a learning rate of $2\times10^{-6}$.

\textbf{Self-Forcing DMD Distillation.} 
In this stage, we use a batch size of 32 and train for approximately 2k steps. The learning rates are set to $2\times10^{-6}$ for the generator and $4\times10^{-7}$ for the critic (i.e., the fake score function), with an update ratio of 1:5 between the generator and the critic.

\section{Details on DisCoBench and Evaluation}\label{appendix: A}
\textbf{Design of the DisCo-ActionBench.}
DisCo-ActionBench is constructed to evaluate models under a wide range of complicated camera action sequences rather than repeated instances of a small number of simple trajectories. Each sample is associated with a structured camera motion sequence composed of discrete motion primitives from our predefined action set $\mathcal{A}$, and contains at least two types of actions.
Each camera action sequence can be formulated as
$seq = (a_1 \times N_1)(a_2 \times N_2)...$
where $a_i \in \mathcal{A}$ and $N_i$ represents the lasting number of action $a_i$. To avoid changing actions too quickly, we set $N{i} \geq 4$.

\begin{figure}[!htbp]
\vskip 0.1in
  \begin{center}
    \centerline{\includegraphics[width=0.4\columnwidth]{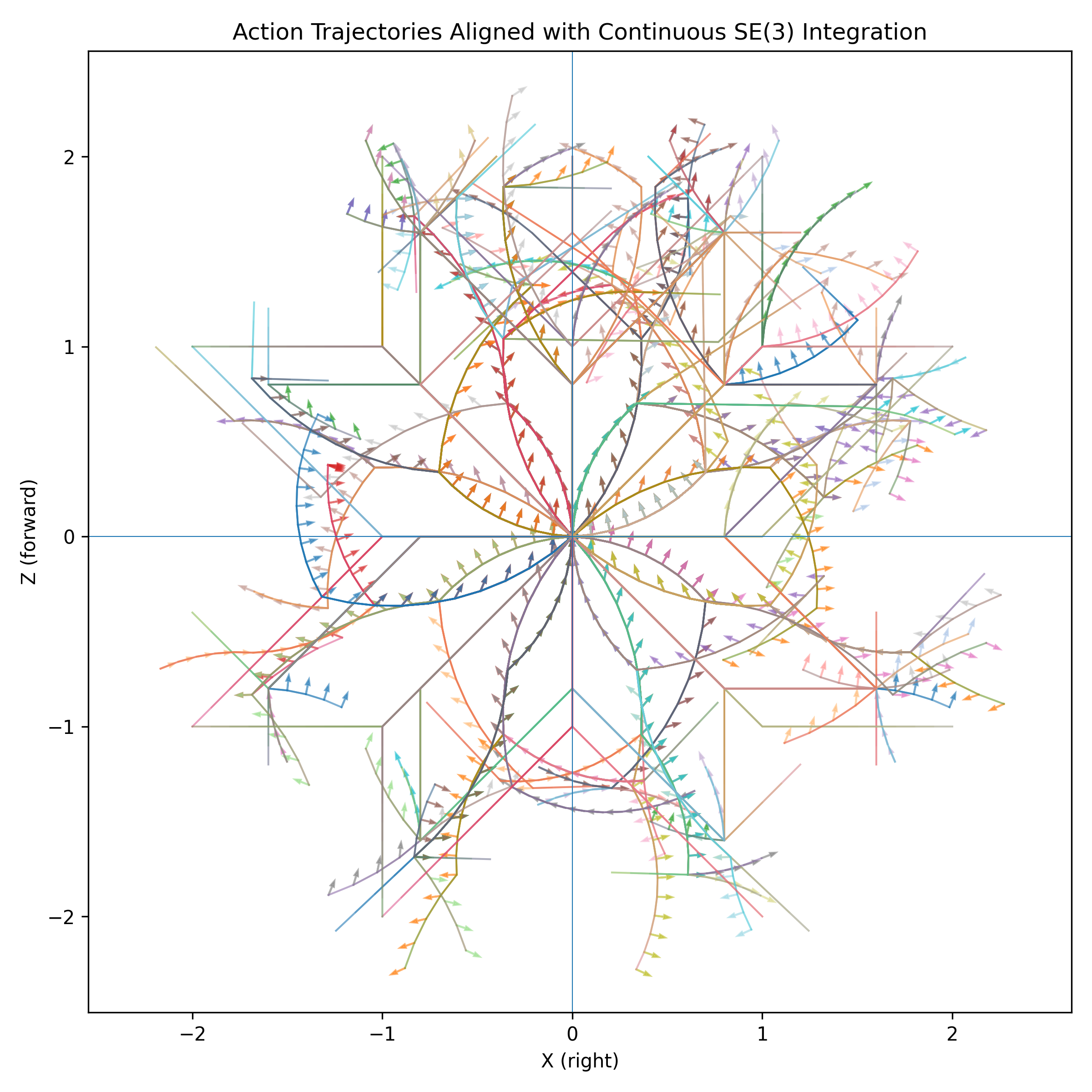}}
    \caption{
    Each polyline corresponds to a unique camera motion sequence in the DisCo-ActionBench. All trajectories start from the same origin, and are constructed by sequentially applying discrete camera motion primitives. The visualization highlights the diversity of trajectory structures and directional coverage.
    }
    \label{visualization_action_trajs}
  \end{center}
\end{figure}

We also provide a visualization to better present the trajectories. As illustrated in \ref{visualization_action_trajs}, DisCo-ActionBench covers a wide
range of action sequences structures, offering a strong evaluation suite for the assessment of the controllability of models.

\textbf{Effectiveness of VQA-Score in DisCo-DynamicBench.}
In this section, we provide additional details to demonstrate the effectiveness of the VQA-Score used in DisCo-DynamicBench. Since most samples in DisCo-DynamicBench are sourced from OmniWorld-Game~\cite{omniworld}, we directly adopt the captions provided by the dataset for VQA-Score calculation. For the remaining samples, we follow the same caption generation protocol as OmniWorld-Game to produce corresponding textual descriptions.

\begin{figure}[t]
  \vskip 0.1in
  \begin{center}
    \centerline{\includegraphics[width=\columnwidth]{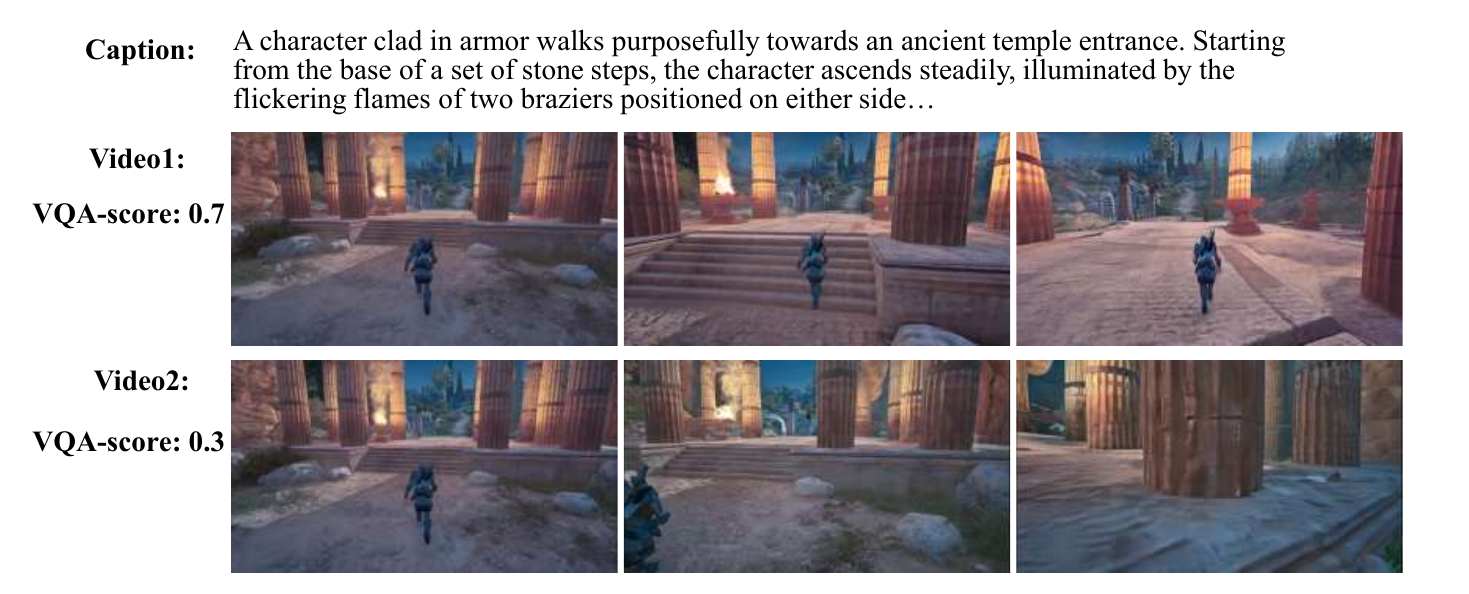}}
    \caption{
    Comparison of two generated videos under the same caption with their corresponding VQA-Scores.
    Video1 aligns well with the caption by exhibiting the required dynamic content and achieves a higher VQA-Score, whereas video2 mainly contains camera motion in a static scene and receives a significantly lower score.
    }
    \label{vis_vqa}
  \end{center}
\end{figure}

As shown in Fig.\ref{vis_vqa}, since the captions explicitly describe the dynamic content expected in the generated videos, VQA-Score is able to effectively assess whether the generated videos contain the required dynamic elements, rather than merely exhibiting camera motion in static scenes. 
\section{More Experiments and Discussions}
\subsection{Sensitivity Analysis of Action Score}
\begin{figure}[!htbp]
  % \vskip 0.2in
  \begin{center}
    \centerline{\includegraphics[width=\textwidth]{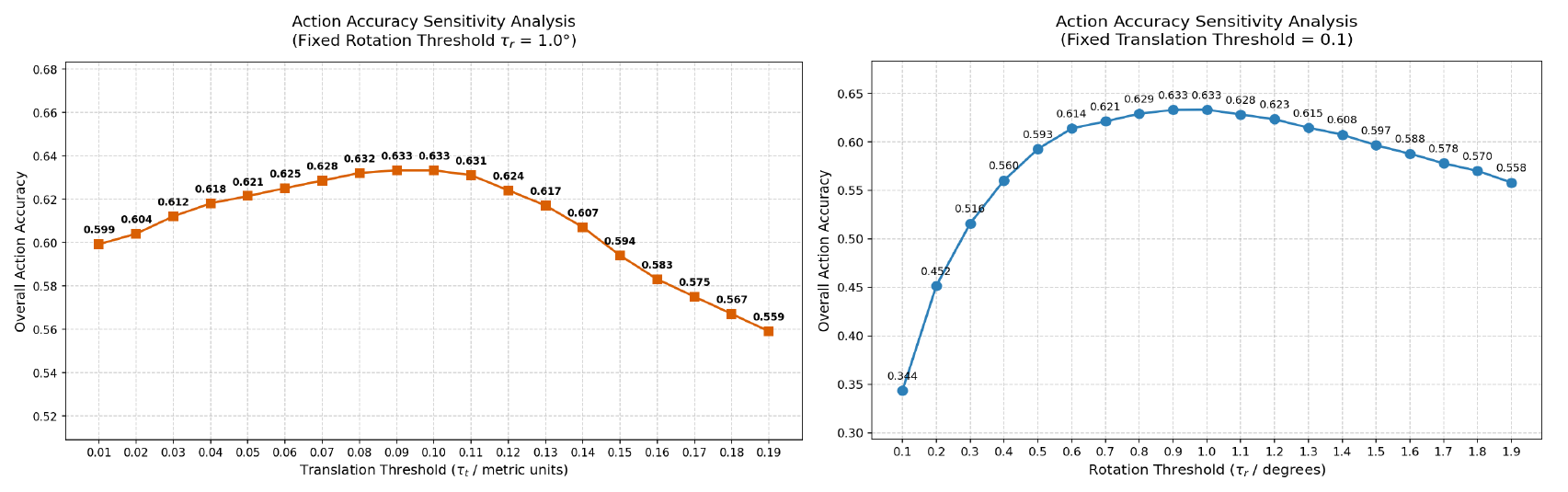}}
    \caption{
Sensitivity Analysis of Action Score
    }
    \label{fig:sensitivity}
  \end{center}
\end{figure}

To examine the sensitivity of the action score, we vary the translation threshold $\tau_t$ and rotation threshold $\tau_r$ and report the results in Figure~\ref{fig:sensitivity}.

We find that the action score remains stable across a reasonably broad range of thresholds: it stays above $0.61$ for $\tau_r \in [0.6, 1.3]$ and above $0.62$ for $\tau_t \in [0.05, 0.12]$. This demonstrates the robustness of the metric and the validity of our chosen settings. 
\subsection{Trade-off between Discretization and Fine-grained Control}
Discretizing continuous trajectories into 27 bins inevitably introduces some fine-grained information loss. However, continuous camera trajectories, though fully fine-grained, often lead to feature entanglement and degraded controllability. Our discretized actions instead trade off sub-bin precision for a more structured and distinguishable action space, which we validate is fundamental to fine-grained control as complex motions can only be modeled reliably when different actions can be represented in a sufficiently separable manner.

We further support this claim with an additional ablation on the number of bins. As shown in Table~\ref{ablation_on_bins}, increasing the number of bins indeed provides a finer-grained action space, but it also leads to worse Action Score. This result further validates our argument that stronger expressiveness alone does not necessarily improve controllability; when the action space becomes overly fine-grained and less distinguishable, control fidelity can instead degrade. 
\begin{table}[!htbp]
  \caption{Effect of number of bins.}
  \label{ablation_on_bins}
  \begin{center}
    \begin{small}
      \setlength{\tabcolsep}{14pt}
      \begin{tabular}{l c}
        \toprule
        Number of Bins & Action Score \\
        \midrule
        27 (9$\times$3)  & \textbf{0.6332} \\
        39 (13$\times$3) & 0.5638 \\
        48 (16$\times$3) & 0.5073 \\
        \bottomrule
      \end{tabular}
    \end{small}
  \end{center}
\vspace{-1em}
\end{table}
\subsection{Human Evaluation} \label{app: huamn_eval}
To mitigate the concern that automatic evaluators may introduce bias, we additionally conduct an independent human evaluation that does not rely on VIPE or symbolic discretization. Specifically, we recruit 30 human evaluators(consisting of 15 researchers with relevant background in computer vision or machine learning, and 15 non-expert participants), and present them with pairs of videos generated by different methods under the same action condition. For each pair, evaluators are asked to make two separate judgments: (1) which video better follows the given action condition, and (2) which video exhibits smoother and more natural motion. The presentation order of videos within each pair is randomized to avoid positional bias.

We aggregate the pairwise preferences separately for each criterion across all evaluators. This allows us to independently assess performance along these two dimensions and demonstrate the advantage of our model in both action adherence and motion quality. As shown in Table~\ref{huamn_evalution}, our model is consistently preferred over baselines.
\begin{table}[!htbp]
  \caption{Human evaluation results.}
  \label{huamn_evalution}
  \begin{center}
    \begin{small}
      \setlength{\tabcolsep}{16pt}
      \begin{tabular}{lcc}
        \toprule
        Method & \makecell{Action \\ Following} & \makecell{Motion \\ Smoothness} \\
        \midrule
        Ours & \textbf{65.5\%} & \textbf{73.7\%} \\
        Baselines & 34.5\% & 26.3\% \\
        \bottomrule
      \end{tabular}
    \end{small}
  \end{center}
\end{table}
\subsection{Generalization Ability}
To evaluate the generalization capability of DisCo, we randomly select 100 videos from the external Sekai~\cite{sekai} dataset, and compare our method with Matrix-Game2. As shown in Table~\ref{generalization}, our model also exhibits competitive performance on unseen real-world data.
\begin{table}[!htbp]
  \caption{Performance on unseen real-world data.}
  \label{generalization}
  \begin{center}
    \begin{small}
      \setlength{\tabcolsep}{16pt}
      \begin{tabular}{lcc}
        \toprule
        Method & FVD $\downarrow$ & Action Score $\uparrow$ \\
        \midrule
        Ours & \textbf{333.93} & \textbf{0.602} \\
        Matrix-Game2 & 369.10 & 0.567 \\
        \bottomrule
      \end{tabular}
    \end{small}
  \end{center}
\end{table}
\subsection{Multi-seed Results} \label{multi-seed section}
To further validate that our gain in controllability is reproducible rather than due to randomness, we evaluate our model over four independent runs with different random seeds on DisCo-ActionBench. As shown in Table~\ref{multi_seed}, the Action Score remains stable with low standard deviations, which reflects genuinely better controllability.
\begin{table}[!htbp]
  \caption{Multi-seed results.}
  \label{multi_seed}
  \begin{center}
    \begin{small}
      \setlength{\tabcolsep}{12pt}
      \begin{tabular}{lccccc}
        \toprule
        Method & score-r1 & score-r2 & score-r3 & score-r4 & Std. \\
        \midrule
        Ours & \textbf{0.6332} & \textbf{0.6315} & \textbf{0.6253} & \textbf{0.6357} & \textbf{0.0044} \\
        Matrix-Game2 & 0.5880 & 0.5863 & 0.5912 & 0.5800 & 0.0047 \\
        \bottomrule
      \end{tabular}
    \end{small}
  \end{center}
\end{table}
\subsection{Runtime Characteristics}
The runtime characteristics of our method and the comparison with self-forcing are reported in Table~\ref{efficiency}.Our method is more resource expensive mainly for its larger backbone and higher supporting resolution. This evaluation is conducted on a single 48G 4090.
\begin{table}[!htbp]
  \caption{Efficiency comparison.}
  \label{efficiency}
  \begin{center}
    \begin{small}
      \setlength{\tabcolsep}{10pt}
      \begin{tabular}{lcccc}
        \toprule
        Method & Frame-rate & Latency & Resolution & Memory \\
        \midrule
        Ours (5B) & 16 & 23.4s & 1280$\times$704 & 33G \\
        Self-forcing (1.3B) & 16 & 13.8s & 832$\times$480 & 16G \\
        \bottomrule
      \end{tabular}
    \end{small}
  \end{center}
\end{table}
\section{Limitation} \label{limitations}
Although \system covers most common actions for user interactions, including basic translational movements, diagonal movements, yaw rotations, and their combinations, some actions such as pitch, roll, and variable velocities are currently not supported. This limitation stems mainly from the fact that constructing large-scale, action-balanced datasets that cover all types of action patterns is highly resource-intensive. We leave this to future work. 
Another limitation arises in applications that require very precise camera placement, as the discretization process inevitably sacrifices some fine-grained information and expressiveness. However, we observe that overly fine-grained action spaces (e.g., continuous actions) lead to feature entanglement and degraded controllability. Therefore, we trade off sub-bin precision for improved controllability.

\section{Demos}\label{appendix: demo}

\begin{figure}[!htbp]
  \vskip -0.15in
  \begin{center}
    \centerline{\includegraphics[width=0.9\textwidth]{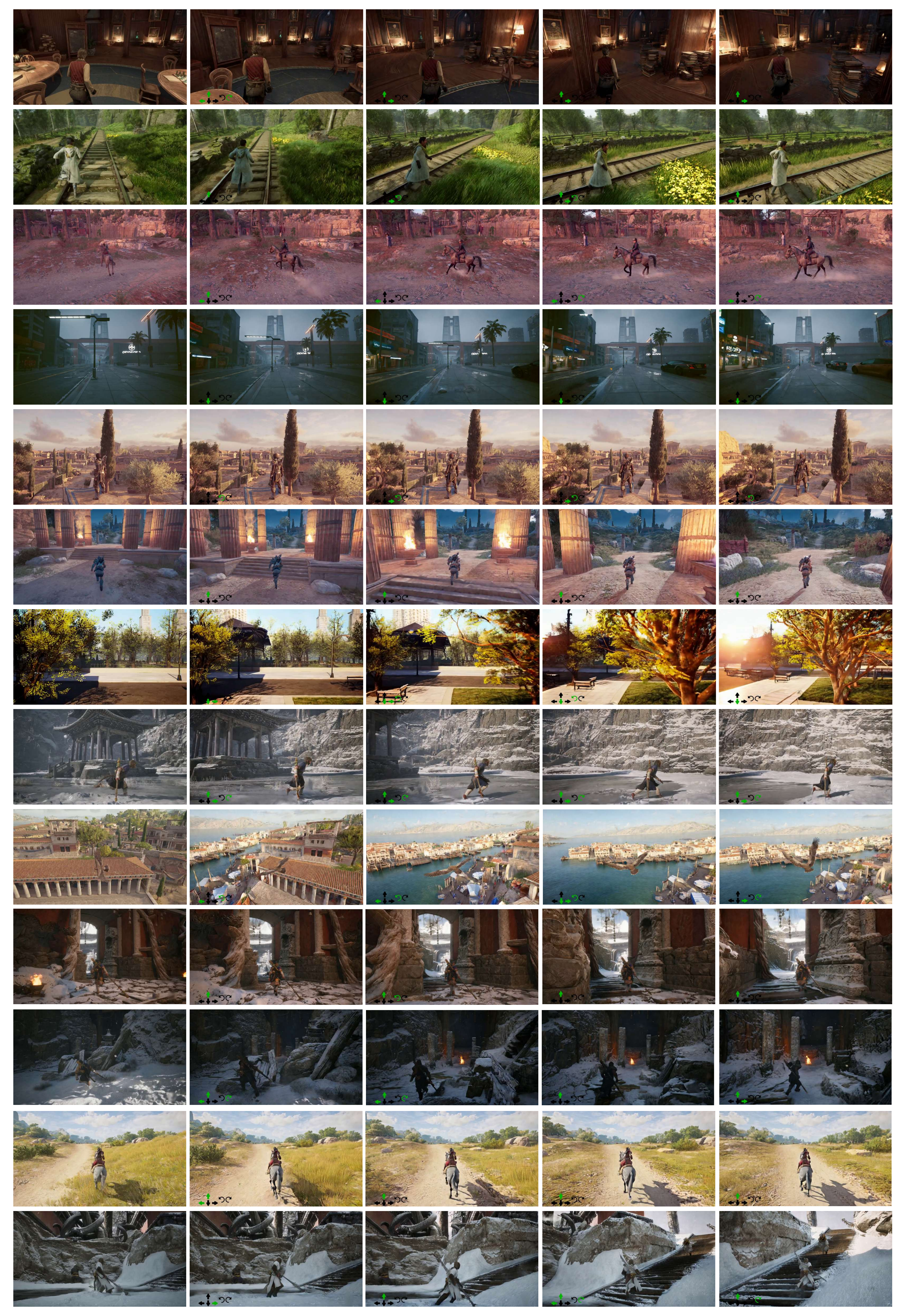}}
    \label{demo2}
    \caption{
    Demos
    }
  \end{center}
\end{figure}